\documentclass[letterpaper, 10 pt, conference]{ieeeconf}  % Comment this line out if you need a4paper

\IEEEoverridecommandlockouts                              % This command is only needed if
\usepackage{algorithm}
\usepackage{algorithmic}
\usepackage{graphicx}
\usepackage{subfigure}
\usepackage{algorithm}
\usepackage{algorithmic}
\usepackage{graphicx}
\usepackage{subfigure}
\usepackage{amsmath}
\usepackage{bbm}
\usepackage{amsfonts,amssymb}
\usepackage{setspace}
\usepackage{booktabs}
\usepackage{color}
\usepackage{lineno}
\usepackage{newfloat}
\usepackage{listings}
\usepackage[marginal]{footmisc}
\usepackage{amsfonts,amssymb}
\usepackage{url}
\usepackage{dsfont}

\title{\LARGE \bf
Reward-Adaptive Reinforcement Learning: Dynamic Policy Gradient Optimization for Bipedal Locomotion
}
\author{Changxin Huang$^{1}$, Guangrun Wang$^{2}$, Zhibo Zhou$^{3}$, Ronghui Zhang$^{1}$ and Liang Lin$^{3}$% <-this % stops a space
\thanks{$^{1}$Changxin Huang and Ronghui Zhang are with the School of Intelligent
Systems Engineering, Sun Yat-sen University, Guangzhou 510275, China.}%
\thanks{$^{2}$Guangrun Wang is with the Department of Engineering Science, University of Oxford, UK.}
\thanks{$^{3}$Zhibo Zhou and Liang Lin are with the School of Computer Science and Engineering, Sun Yat-sen University, Guangzhou 510275, China.}%
}

\begin{document}

\maketitle
\thispagestyle{empty}
\pagestyle{empty}

%%%%%%%%%%%%%%%%%%%%%%%%%%%%%%%%%%%%%%%%%%%%%%%%%%%%%%%%%%%%%%%%%%%%%%%%%%%%%%%%
\begin{abstract}

Controlling a non-statically bipedal robot is challenging due to the complex dynamics and multi-criterion optimization involved. Recent works have demonstrated the effectiveness of deep reinforcement learning (DRL) for simulation and physical robots. In these methods, the rewards from different criteria are normally summed to learn a single value function. However, this may cause the loss of dependency information between hybrid rewards and lead to a sub-optimal policy. In this work, we propose a novel reward-adaptive reinforcement learning for biped locomotion, allowing the control policy to be simultaneously optimized by multiple criteria using a dynamic mechanism. The proposed method applies a multi-head critic to learn a separate value function for each reward component. This leads to hybrid policy gradient. We further propose dynamic weight, allowing each component to optimize the policy with different priorities. This hybrid and dynamic policy gradient (HDPG) design makes the agent learn more efficiently. We show that the proposed method outperforms summed-up-reward approaches and is able to transfer to physical robots. The sim-to-real and MuJoCo results further demonstrate the effectiveness and generalization of HDPG.

\end{abstract}

%%%%%%%%%%%%%%%%%%%%%%%%%%%%%%%%%%%%%%%%%%%%%%%%%%%%%%%%%%%%%%%%%%%%%%%%%%%%%%%%
\section{Introduction}

Bipedal robots locomotion  \cite{hereid2018dynamic} is about controlling a robot to walking on different surfaces with stable walking gaits. This task is challenging because of the complex dynamics involved especially where undulated surfaces or obstacles are present. Traditional hand-crafted control policies, such as zero moment point (ZMP) \cite{kajita2014introduction} and hybrid zero dynamics (HZD) \cite{ames2016dynamic} and divergent component of motion (DCM) \cite{pratt2012capturability}, often suffer from limited adaptation to various environments.

%Therefore, Artificial-Neural-Network(ANN)-based methods were proposed and target this problem as an imitation problem, in which a control policy manually designed by human experts acts as a golden standard and a supervised learning strategy is adopted to approximate the learning results to this reference policy \cite{xie2020learning,imi1,dart,siekmann2020learning}. These methods still rely on human expertise for the reference policy design and hence can be laborious and tricky to repeat the results.

% This task becomes especially challenging when robots are deployed to complicated environments where random obstacles and undulated ground surfaces are present.
% Meanwhile, it can be difficult for manually planned control policies to handle above situations.
% Many existing ANN-based methods modelling this problem as an imitation problem, in which a control policy manually designed by human experts acts as a golden standard and a supervised learning strategy aims to approximate the learning results to this reference policy \cite{xie2020learning,imi1,dart,siekmann2020learning}. These methods still rely on human expertise for the reference policy design and hence can be laborious and tricky to repeat the results.
Recently, reinforcement learning (RL) has made significant progress in solving complex biped locomotion problems \cite{siekmann2020learning, xie2020learning, castillo2019hybrid, li2019using}. The core of RL is training robots to take actions that maximize the expected cumulative rewards. A reward function usually consists of multiple components \cite{zhang2019teach, xie2018feedback}, each of which quantitatively describes an aspect of the quality for the walking task, such as body balance maintenance, limb-alteration gait, conservative motor torques and so on. Most existing approaches simply add up the component rewards to learn a single value function \cite{siekmann2020learning, xie2018feedback}, which may break the correlation between different rewards and therefore limit learning efficiency \cite{van2017hybrid, flet2019merl}.

\begin{figure}[]
\centering
\includegraphics[width=0.8\columnwidth]{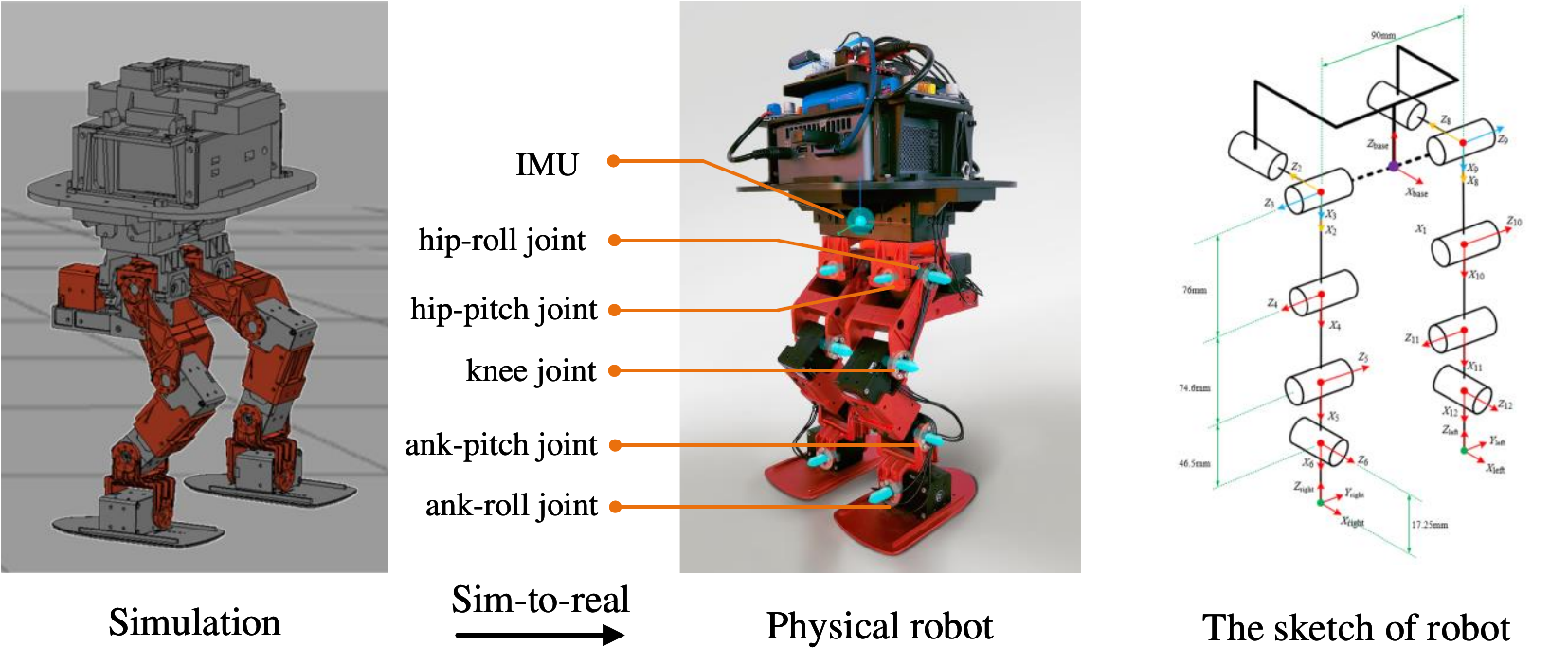} % Reduce the figure size so that it is slightly narrower than the column. Don't use precise values for figure width.This setup will avoid overfull boxes.
\caption{\textbf{Left}: The simulation bipedal robot. \textbf{Middle}: The physical bipedal robot. \textbf{Right}: The sketch of robot.}
\label{fig:robot}
\end{figure}

In this work, we propose a hybrid and dynamic policy gradient (HDPG) optimization method to address the above-mentioned issues. Specifically, we construct a multi-head critic in the deep deterministic policy gradient (DDPG) \cite{LillicrapHPHETS15} framework to capture the gradients separately obtained from multiple rewards. Each head exclusively learns from corresponding reward feedback. %(see Figure \ref{framework}).
Meanwhile, the branch gradients are merged during back-propagation with dynamically updating weights, which aims to guide the policy network to first learn from the ``simple" components before the ``challenging" ones. The motivation for this dynamic mechanism is that we tentatively guide the robot to give higher priority to learn from components that show fast reward accumulation. For example, the agent is encouraged to learn to maintain body balance before learning to move forward. Intuitively, keeping balance is the prerequisite of moving forward. Thus, the body balance component should have a higher priority than moving forward in the initial learning stage. To achieve this goal, we introduce dynamic weights to allow each component gradient to optimize the policy with a different priority. Therefore, the HDPG agent adaptively learns each reward component during training. The contributions of this work are summaries as following:

\begin{itemize}
	\item We introduce a multi-head critic to learn a separate value function for each component reward. The experimental results show that our proposed multi-head design outperforms the traditional reward-sum method in biped locomotion tasks.
	
	\item We propose dynamic weights for hybrid policy gradients to improve learning efficiency. In this way, the control policy is optimized by hybrid policy gradients in a dynamic manner.
	
	\item We build our bipedal robot in the gazebo simulator. The sketch and the corresponding physical robot are shown in Fig. \ref{fig:robot}. We also release 3 challenges, which are push recovery challenges, obstacle challenges and slope terrain challenges. We train HDPG policy with simple dynamics randomization in the simulator. The policy is successfully transferred to the physical robot without further tuning.
	
    \item We further conduct experiments on OpenAI gym \cite{brockman2016openai} to verify the generalization of HDPG. Experimental results demonstrate that the proposed method can be applied to more general continuous control tasks to improve learning efficiency.

\end{itemize}

\section{Related work}

Bipedal locomotion problems were mostly targeted by manually designed polices \cite{ponton2021efficient, carpentier2018multicontact, dai2016planning} using methods such as zero moment point (ZMP) \cite{kajita2003biped, kajita2014introduction}, hybrid zero dynamics (HZD)\cite{castillo2020hybrid, ames2016dynamic, ames2014rapidly}, divergent component of motion (DCM) \cite{englsberger2015three, pratt2012capturability} and so on. ZMP-based methods \cite{kajita2003biped, kajita2014introduction} use the simplified models, e.g., linear inverted pendulum(LIP)dynamics, to generate walking patterns with the constraints that ZMP should be maintained in the support polygon. DCM-based methods \cite{englsberger2015three, pratt2012capturability} simplify gait generation using the divergent component of center of mass (CoM) which enlarges the support polygon and allows the robot recover from external perturbations by step adjustment. HZD approaches \cite{ames2016dynamic,ames2014rapidly} impose virtual constraints to put the biped into zero dynamics surface via feedback linearization. However, the flexibility of these manually designed polices is limited, therefore, it is challenging to handle difficult situations where obstacles and inclined surfaces are present. In addition, such manually designed policies heavily based on the precondition that the robot dynamics and kinematic are known and determined. They cannot be implemented on robots that dynamics and kinematic are unknown.

To generalize the walking capability for bipeds to such challenging situations and enable the agent to learn the policy without knowing robot dynamics, imitation-learning (IL) based approaches aim to train neural network models supervised by expert policies \cite{xie2020learning,imi1,dart,siekmann2020learning}. But these methods rely on human experts to manually design a control policy as reference, which can be laborious.

%Many approaches for bipedal locomotion use simplified dynamic model to avoid sophisticated dynamics. Those approaches like LIPM \cite{kajita2003biped}\cite{kajita2014introduction}  or SLIP \cite{xiong2018coupling} are simple enough to be implemented on real time control but result in too many physical constrains. Trajectory optimization approach  with tracking feedback controller is also proposed to utilize the dynamics of high-dimension model. However, these approaches generate simple periodical gaits for robot. They are always limited to some specific terrains or scenes which means it takes many efforts to transfer the robot from one to another.

Reinforcement learning system optimizes the policy by exploration and interaction with a simulation environment. It encourages favorable movements and punishes improper ones to learn an optimal policy. In order to address the aforementioned issues of traditional controller and imitation learning, many researches attempt to learn bipedal locomotion policies following a reinforcement learning framework \cite{peng2017deeploco, yu2018learning}. However, they did not demonstrate the effectiveness on physical robots. Recently, Xie \emph{et al.} \cite{xie2018feedback} formulates a feedback control problem as finding the optimal policy for a Markov Decision Process to learn robust walking controllers that imitate a reference motion with DRL. This proved to be effective on Cassie simulation. Subsequently, they use an iterative training method and Deterministic Action Stochastic State (DASS) tuples to learn a more robust policy \cite{xie2020learning}. Siekmann \emph{et al.} \cite{siekmann2020learning} introduce recurrent neural networks (RNNs) to learn memory features that reflect physical properties. However, all of these approaches use the sum of rewards to learn a single value function, which may cause the loss of the dependency information between different rewards and limit the learning efficiency of value function \cite{van2017hybrid}.

In this paper, we aim to make full use of the dependency between different reward components to improve learning efficiency, which is related to Hybrid Reward Architecture (HRA) \cite{van2017hybrid}, Decomposed Reward Q-Learning (drQ) \cite{juozapaitis2019explainable}, and DDPG \cite{LillicrapHPHETS15}. Specifically, we first proposed a multi-head critic to learn a separate value function for each component reward function, which is similar to HRA \cite{van2017hybrid} and drQ \cite{juozapaitis2019explainable}. %and Multi-Head Reinforcement Learning (MERL) \cite{flet2019merl}.
HRA and drQ first decompose the reward function of the environment into $n$ different reward functions. They aim to learn a separate value function; each of them is assigned a reward component. Learning a separate value function is proved to enable more effective learning. However, both HRA and drQ are based on Q-learning and can only be used for discrete action space tasks. Our HDPG is based on DDPG \cite{LillicrapHPHETS15}, which allows learning continuous actions. Moreover, we further propose the dynamic weight for hybrid policy gradients to optimize the policy with different priorities. This hybrid and dynamic policy gradient (HDPG) design make the agent learn more efficiently.

\section{Preliminaries}

In this section, we will briefly introduce the the background and annotations of RL and the theory of deep deterministic policy gradient (DDPG) \cite{LillicrapHPHETS15}, which is the base model of the proposed method.

\subsection{Markov Decision Process(MDP)}

Specifically, biped locomotion is formulated as a Markov Decision Process problem. A standard reinforcement learning includes an agent interacting with an environment $E$ and receiving a reward $r$ at every time step $t$ in MDP. MDP can be denoted as the tuple $(\mathcal{S},\mathcal{A},\mathcal{P},\mathcal{R},\gamma)$, where $\mathcal{S}$ is the state space, $\mathcal{A}$ is the action space, {blue} $\mathcal{P}:\mathcal{S} \times \mathcal{A} \mapsto \mathcal{S}$ is the transition probability function, and $\mathcal{R}$ is the reward function, $\gamma \in (0,1]$ is the discount factor.

At every time step $t$, the robot observes current state $s_{t} \in \mathcal{S}$, and takes an action $a_t$ according to a policy $\pi:a_t = \pi(s_t)$. Then, the agent transits to the next state $s_{t+1} = \mathcal{P}(s_{t+1}|s_t,a_t)$ and receives a reward {${r_t}=\mathcal{R}(s_{t},a_{t})$} as well as a new state $s_{t+1} \in \mathcal{S}$ from the environment. The return from a state $s_t$ as the cumulative $\gamma$-discounted reward: {$\sum\nolimits_{k = t}^T {{\gamma ^{k - t}} \mathcal{R}(s_k,a_k)}$}.
The objective of RL is to optimize the policy $\pi$ by maximizing the expected return from the initial state.
According to the \emph{Bellman Equation}, the expected return starts from state $s_t$, takes action $a_t$, and follows policy $\pi$, which is denoted as action-value function ${Q_\pi}({s_t},{a_t})$:
%The action-value function ${Q_\pi}({s_t},{a_t})$, \emph{i.e.}, $Q$-function, is the expected return starting from state $s_t$, taking action $a_t$, following policy $\pi$ according to the \emph{Bellman Equation}:
\begin{equation}
\small
{Q_\pi }({s_t},{a_t}) = {{\mathds E}_{{r_t},{s_{t + 1}} \sim E}}[r({s_t},{a_t}) + \gamma {Q_\pi }({s_{t + 1}},\pi ({s_{t + 1}}))]
\label{Q_function}
\end{equation}

\subsection{Deep Deterministic Policy Gradient}
Deep Deterministic Policy Gradient (DDPG) \cite{LillicrapHPHETS15} is a representative model-free RL method. Considering its remarkable performance for the continuous control problem, DDPG is utilized as our basic model. It consists of a $Q$-function (the critic) and a policy function (the actor) to learn a deterministic continuous policy.
The actor $\pi$ and the critic $Q$ are learnt with deep function optimization, which is parameterized by $\theta^\pi$ and $\theta^Q$, respectively.

To avoid the agents falling into a local optimum and further improve the exploration efficiency, an exploration noise is introduced to the policy. The output action can be expresses as: ${a_t} = \pi ({s_t}) + {\mathcal{N}_t}$, where $\mathcal{N}_t \in \mathcal{N}_{OU}$. %The set refer to the code from Ben Lau \footnote{\emph{\url{https://yanpanlau.github.io/2016/10/11/Torcs-Keras.html}}}: $\mathcal{N}_t=\lambda \times (\mu-x)+\sigma$, where $\lambda$ means the how fast the variable reverts towards to the mean£¬ $x$ denotes the current action, $\mu$ represents the mean value. $\sigma$ is a random variable sampled from standard normal distribution.
It utilizes the experience replay strategy which collects the experience tuples $(s_t, a_t, r_t, s_{t+1})$ stored in a replay buffer $B$ during exploration. In training stage, the training data are sampled from the replay buffer to optimize the policy. DDPG approximates $Q$-function by a critic network, which is updated by minimizing the \emph{Bellman} loss:
\begin{equation}
{L}({\theta ^Q}) = {\mathds E}_{s_t, a_t, r_t, s_{t+1} \sim B} ({y_t} - {Q}({s_t},{a_t}|{\theta ^Q})),
\label{loss_original}
\end{equation}
where ${y_t} = {r_t} + \gamma {Q}({s_{t + 1}},\pi({s_{t + 1}})|{\theta ^{Q}})$ is the target value. Then the actor learns a $Q$-optimal policy: $\pi(s) = argmax{_a}Q(s,a)$ and optimizes the policy by using the sampled policy gradient:
%%%%%%%%%%%%%%%%%%%%%%%%%%%%%%%%%%%%%%%%%%%%%%%%%%%%%%
\begin{equation}
\small
{\nabla _{{\theta ^\pi }}}J \approx
{\mathds E}_{s_t, a_t \sim B}[{\nabla _a}Q(s_t,\pi(s_t)|{\theta ^Q}){\nabla _{{\theta ^\pi }}}\pi (s_t|{\theta ^\pi })].
\label{J_original}
\end{equation}

%For stabilize training, DDPG introduced a target actor network $\theta^{\pi{'}}$ and a target critic network $\theta^{Q{'}}$ respectively, which are used for calculating the target values. The target network is updated through soft update \cite{LillicrapHPHETS15}: $\theta{'} = \tau \theta + (1-\tau)\theta{'}$, where $\tau$ is the update rate.

\begin{figure}[t]
\centering
\includegraphics[width=0.85 \columnwidth]{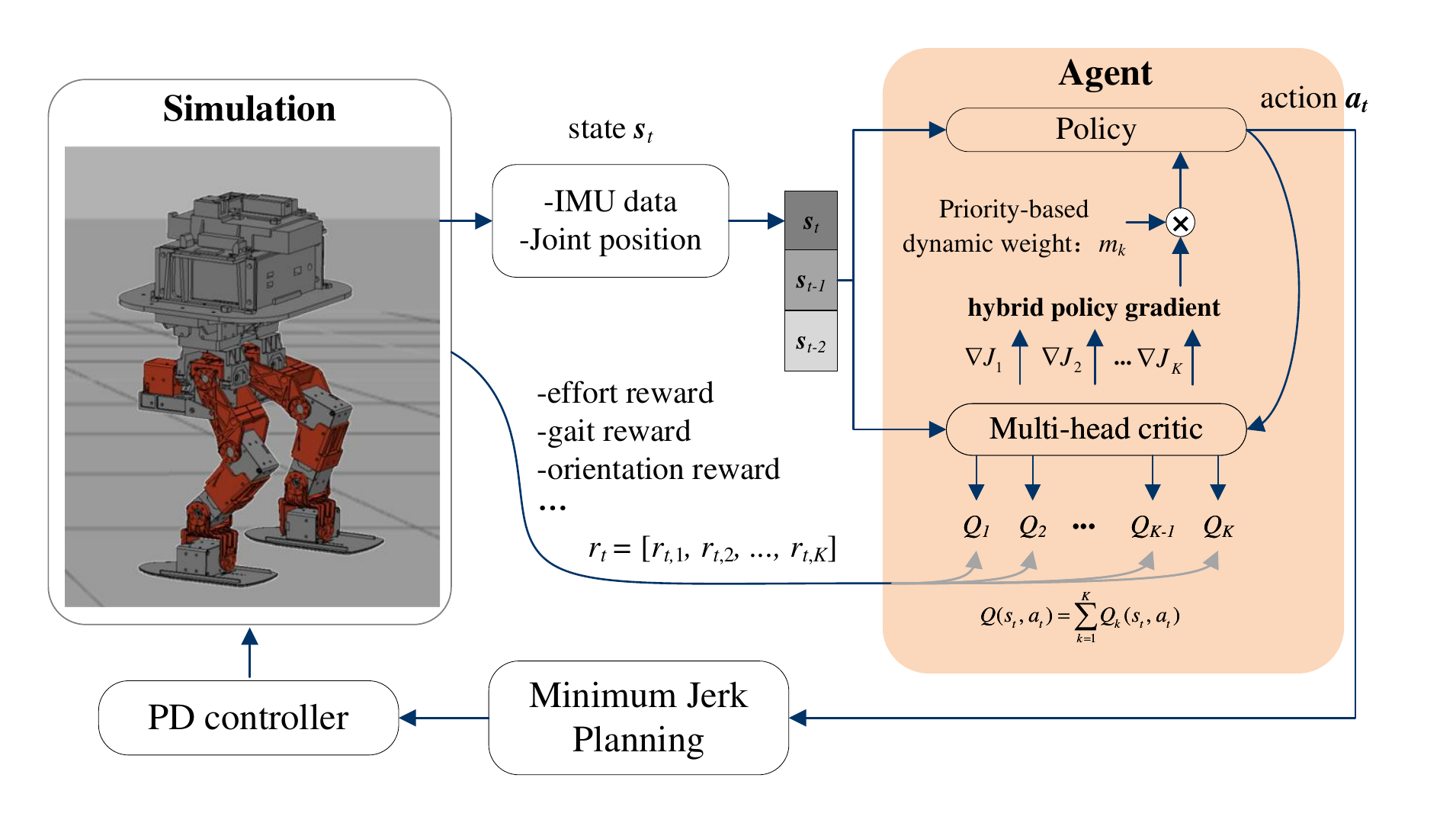} % Reduce the figure size so that it is slightly narrower than the column. Don't use precise values for figure width.This setup will avoid overfull boxes.
\caption{Overview. The biped robot interacts with the simulator and obtains experience transitions $(s_t, a_t, s_{t+1}, r_t)$. The multi-head critic of HDPG learns a separate $Q$-value function for each reward component. The dynamic weight is assigned for the hybrid policy gradient to adjust the learning priority of each policy gradient component.}
\label{framework}
\vspace{-15pt}
\end{figure}

\section{Method}
\label{sec:method}
%In this section, we will firstly provide an overview of the proposed Progressive reinforcement learning (PRL) for biped locomotion. %The details of learning a multi-head critic are elaborated in Sec.\ref{sec:multi-head}. In order to make the robot learn gait faster, simple regular planning is introduced in Sec. \ref{sec:hip-planning}, which stimulates the robot visit more gait states. Finally, we describe the reward function in Sec.\ref{sec:hip-planning}.

The framework of our method is given in Fig. \ref{framework}. The robot observes the current state $s_t$ from Gazebo environment, which contains the data from the inertial measurement unit (IMU) and angular positions of all joints. Inspired by ``memory-based control" \cite{siekmann2020learning, zhang2019teach}, we use state sequence $(s_{t-2}, s_{t-1}, s_{t})$ as the input of actor to contain temporal information. The predict action $a_t$ from the actor contains the control signal about the angular position of joints. It is sent to a minimum jerk planning module \cite{kyriakopoulos1988minimum} to generate their smooth transition trajectories. A low level PD controller is applied to track these joint position trajectories. After that, the agent will receive a reward $r_t$ from the environment. This process is iterative until termination.

Learning biped locomotion is a complex task due to multiple constraints involved \cite{gil2019learning, xi2019balance, wu2018motion}. When walking forward, the robot needs to maintain the balance of the pelvis and avoid the abnormal states, such as falling down, over-torque and strange posture. This usually results in multi-dimensional rewards, each of which corresponds to a constraint. Existing methods usually use the weighted sum of rewards to learn a single value \cite{zhang2019teach, xie2020learning, siekmann2020learning}: $r_t= \sum\nolimits_{k = 1}^K {w_k{r_{t,k}}}$, where $K$ is the number of rewards and $w_k$ is the weight of each reward. However, simply summing the rewards will lose the dependency between different rewards. Two sets of very different rewards may get the same sum, which results in $Q$-value not giving insight into factors contributing to policy.
For example, when the robot receives a negative reward, it cannot understand whether it is due to the unnatural pose or the over-torque of servomotors.

\subsection{Multi-head Critic for multiple values learning} \label{sec:multi-head}

Inspired by ``hybrid reward architecture (HRA)" \cite{van2017hybrid}, we conduct a multi-head critic to learn a separate value function for each component reward function, as shown in Fig. \ref{framework}. Therefore, the reward is decomposed and can be expressed as a vector: $r_t = [r_{t,1}, r_{t,2}, ..., r_{t,K}]$. Each head of the critic learns an action-value corresponding to a specific sub-reward. In this way, the overall $Q$-value is also expressed as a vector:
\begin{equation}
\small
Q(s_t,a_t|\theta^Q) = [Q_{t,1},\;Q_{t,2},\;...,\;Q_{t,K}]
\label{Q_vec}
\end{equation}

HRA \cite{van2017hybrid} estimate the $Q$-value based on Deep Q Network (DQN) \cite{mnih2013playing}: $Q_k(s_t, a_t) \leftarrow r_k + \gamma \max\limits_{a_{t+1}} Q_k(s_{t+1}, a_{t+1})$, which can be rewritten as:\begin{equation}\label{eq:hra}
Q_k(s_t, a_t) \leftarrow r_k + \gamma Q_k(s_{t+1}, a_{t+1}(k)),
\end{equation}where $a_{t+1}(k) = \mathop{\arg\max}\limits_{a_{t+1}}Q_k(s_{t+1}, a_{t+1})$. Eq. \eqref{eq:hra} indicates HRA updates different $Q_k$ with different policies (or $a_{t+1}(k)$), ignoring the influence of other components on the overall $Q$-function. To overcome this problem, we adopt a more effective and elegant update in HDPG, i.e., we have: \begin{equation}\label{eq:hdpg}
Q_k(s_t, a_t) \leftarrow r_k + \gamma Q_k(s_{t+1}, a_{t+1}),
\end{equation}where $a_{t+1} = \pi(s_{t+1}|\theta ^ \pi)$. By comparing Eq. \eqref{eq:hra} and \eqref{eq:hdpg}, we can see that our HDPG converges toward the value of the consistent policy. %allowing the policy to produce continuous control.
According to Eq. \eqref{loss_original}, the loss function associated with the multi-head critic is:
% different from HRA and the motivation
% In HRA \cite{van2017hybrid}, the agent adjusts the value of each $Q_k$ toward different policies, which ignores the influence of other objectives. Different from that, in this work, different $y_{t,k}$ are compute with a consistent policy: $\pi(s_{t+1})$.
\begin{equation}
\small
{L}({\theta ^Q}) = {\mathds E}_{s_t, a_t, r_t \sim B} \sum\limits_{k = 1}^K {[{{({y_{t,k}} - {Q_k}({s_t},{a_t}|\theta _k^Q))}^2}]}
\label{loss:multi-head}
\end{equation}
where ${y_{t,k}} = {r_{t,k}} + \gamma {Q_k}({s_{t + 1}},\pi({s_{t + 1}})|{\theta _k^Q})$. Note that we use $\theta _k^Q$ to differentiate the parameters of different $Q_k$. In practical, different $\theta _k^Q$ have share multiple lower-level layers of a critic network. Only the last layer is independent of each other. Then, the overall $Q$-value is defined as the sum of each $Q_k$: $Q({s_t},{a_t}) = \sum\limits_{k = 1}^K {{Q_k}} ({s_t},{a_t})$.

% \begin{figure}[t]
% \centering
% \includegraphics[width=0.5\columnwidth]{./figure/mhcritic.png} % Reduce the figure size so that it is slightly narrower than the column. Don't use precise values for figure width.This setup will avoid overfull boxes.
% \caption{Illustration of the multi-head critic architecture. The multi-head critic learn multiple values, each value corresponding one reward component. Different $Q$-value share the low-level layers parameters.}
% \label{fig:MHC}
% \vspace{-5pt}
% \end{figure}

\subsection{Dynamic policy gradient}\label{sec:progressive}

%According to Eq. \eqref{J_original}, the policy gradient can be re-expressed as:
Eq. \eqref{J_original} can be re-expressed as:
\begin{equation}
\small
\begin{aligned}
{\nabla _{{\theta ^\pi }}}J & \approx {\mathds{E}}_{s_t, a_t \sim B}[{\nabla _a}\sum\limits_{k = 1}^K {{Q_k}(s_t,\pi(s_t)|\theta _k^Q){\nabla _{{\theta ^\pi }}}\pi (s_t|{\theta ^\pi })}] \\
%& =  \sum\limits_{k = 1}^K {{\mathbbm E}_{s_t, a_t \sim B}[{\nabla _a}{Q_k}({s_t},\pi ({s_t})|\theta _k^Q){\nabla _{{\theta ^\pi }}}\pi ({s_t}|{\theta ^\pi })]}\\
& = \sum\limits_{k = 1}^K {\nabla {J_k}}
\end{aligned}
\label{J_hybrid_gradient}
\end{equation}
where ${\nabla {J_k}} = {{\mathds E}_{s_t, a_t \sim B}}[{\nabla _a}{Q_k}({s_t},\pi ({s_t})|\theta _k^Q){\nabla _{{\theta ^\pi }}}\pi ({s_t}|{\theta ^\pi })]$.

This formula indicates that all $Q$-values update the policy with the same weight. In other words, the agent learns all skills in parallel with the same learning rate. However, if there are potential dependencies between reward components, then learning all components with the same learning rate can be challenging and inefficient. At the same time, some components are much easier to learn than others. Therefore, we propose to learn each skill orderly according to the priority. We then evaluate the priority based on the difficulty of each component relative to the current policy.

% Normally, each reward represents the performance of the agent in a certain aspect.
% Specifically, we propose to replace the optimal value function as target for training with an alternative value function that is easier to learn, but still yields a reasonable but generally not optimal policy, when acting greedily with respect to it.
% The key observation behind regularisation on the target function is that two very different value
% functions can result in the same policy when an agent acts greedily with respect to them.

Intuitively, the motivation of dynamic weights is to encourage the agent to learn the easier components first, and then gradually learn the more complex components. To achieve this, the agent first utilizes current policy $\pi_T$ to interact with the environment and obtains $N$ experience samples, which contains $N$ one-step rewards: $r^n = [r_1^n, r_2^n, ..., r_K^n], n\in[1, N] $. All component rewards are normalized to the same value range (e.g., $(0, 1)$). Then, the mean and the variance of reward samples are calculated as $\mu = [\mu_1, \mu_2, ..., \mu_K]$ and $\sigma = [\sigma_1, \sigma_2, ...\sigma_K]$, respectively.

A larger $\mu_k$ means that the $k$ component is easier to get a larger reward, so it is easier to learn for the current policy, and vice versa. The variance reflects the dispersion of each reward's distribution. Stable walking will produce uniform rewards for biped robot, which leads to a small variance $\sigma$. In contrast, the larger variance $\sigma_k$ indicates that the policy is more unstable in component $k$. Intuitively, in order to make the policy more stable, we would like to increase the learning rate of the unstable components.
We define the priority weight $m_k$ of each policy gradient component to mathematically formulate these rules.
%  A smaller variance $\sigma_k$ means that $r_k$ is more sparse, which is more difficult to learn \cite{dann2019deriving}\cite{NIPS2019_9225}\cite{pathak2017curiosity}. In contrast, the larger variance $\sigma_k$ indicates the distribution of samples is more uniform.} Finally, the priority weight $m_k$ of each policy gradient component is defined as follow:

\begin{equation}
\small
{m_k} = \frac{K({{\mu _k} + {e^{\sigma _k^2}}})}{{\sum\limits_{k = 1}^K {({\mu _k} + {e^{\sigma _k^2}})} }}
\label{J_dynamic_weight}
\end{equation}
% \begin{equation}
% {m_k} = \frac{{K{\mu _k}}}{{\sum\limits_{k = 1}^K {{\mu _k}} }}
% \label{J_dynamic_weight}
% \end{equation}
% \begin{equation}
% \nabla J{'_k} = [\beta K{m_k} + (1 - \beta )]\nabla {J_k}
% \label{J_hybrid_dynamic_gradient}
% \end{equation}
Therefore, the total policy gradient is dynamic updated according to: ${\nabla _{{\theta ^\pi }}}J= \sum\limits_{k = 1}^K {m_k \nabla {J_k}}$, where the priority weights will be updated every $T$ episodes.

\subsection{Rewards design}\label{sec:reward shaping}

HDPG does not rely on manually designed reference trajectories to guide the learning process. Existing non-reference methods mainly require careful engineering of the reward function. In other words, the reward function may contain many components (e.g., there are 10 reward components involved in \cite{zhang2019teach}). In this work, only some principles of human gait are utilized to design the reward function. It consists of 6 sub-rewards with regards to walking forward, walking gait, energy consumption and robot pose, which are expressed as step reward, gait reward, height reward, torque reward, orientation reward and fall down reward. The multi-head critic learns the value function with the reward vector, which is defined as: $r_t= [r_t^g\;\;r_t^s\;\;r_t^f\;\;r_t^h\;\;r_t^o\;\;r_t^d]$

\textbf{Gait reward $r_t^g$}: The gait walking is the ultimate goal of the biped robot. The gait occurs when the single support period and double support period arise in time sequence. We formulate this principle as following:
\begin{equation}
r_t^g = \left\{ {\begin{array}{*{20}{l}}
{w_1d_t^g\cos\theta^g,}&{{\rm{if\;a\;gait\;occur}}}\\
{ 0,}&{{\rm{otherwise}}}
\end{array}} \right.
\label{gait_reward}
\end{equation}
where $d_t^g$ is the length of the gait, and $w_1$ is a weight parameter. Assuming $T$ is the time's length of a gait occurs at $s$ time step. $\theta^g$ is the angle between the direction of the robot at $(t-T)$ and $t$. The intention is to encourage the robot to walk forward instead of the other directions.

\textbf{Step reward $r_t^s$}: We also introduce a step reward $r_t^s$ to encourage the robot to move forward at each time step. The step reward function is defined as:
% \begin{figure}[t]
% \centering
% \includegraphics[width=0.68\columnwidth]{./figure/gait.png} % Reduce the figure size so that it is slightly narrower than the column. Don't use precise values for figure width.This setup will avoid overfull boxes.
% \caption{Illustration of a gait. When the robot changes the single support state to the double support state, we define it as a gait.}
% \label{fig_gait}
% \vspace{-10pt}
% \end{figure}
\begin{equation}
r_t^s = w_2d_t^s\cos\theta^s
\label{step_reward}
\end{equation}
where $d_t^s$ is the move distance at $t$ time step, and $w_2$ is a weight parameter of step reward. $\theta^s$ is the angle between the direction of the robot at $(t-1)$ and $t$.

\textbf{Torque reward $r_t^f$}: The torque reward is inverse proportional to the torque magnitude of the joint motors. Negative torque reward is to punish excessive torque of servomotors while walking. The purpose is to lower walking energy and to make walking trajectories smoother. The same torque reward function has proved to be effective in the Robotschool simulator of OpenAI Gym \cite{brockman2016openai}.
\begin{equation}
\small
r_t^t = -w_3\sum\limits_{i = 1}^{I} {|f_t^i|}
\label{force_reward}
\end{equation}
where $I$ is the number of joints, $f_t^i$ is the torque on the $i$-th joint. Meanwhile, the pose of the robot is one of the important indicators for assessing whether the robot is walking normally, which is evaluated with the height and the orientation of the pelvis. The \textbf{height reward} and the \textbf{orientation reward} are defined as:
\begin{equation}
r_t^h = -w_4 |h_t - h_0|
\label{height_reward}
\end{equation}
\begin{equation}
r_t^o = -w_5(|\alpha_t^r|+|\alpha_t^p|)
\label{orientation_reward}
\end{equation}
where $h_t$ is the current height of pelvis and $h_0$ is the desired height. In Equation (\ref{orientation_reward}), $\alpha_t^r$ is the roll angle and $\alpha_t^p$ is the pitch angle of the pelvis. Finally, the \textbf{fall down reward $r_t^d$} is set as -50 for the robot fell to the ground.

\section{Experiments}\label{sec:experiment}

\begin{figure}[t]
\centering
\subfigure[recover from force]{
\begin{minipage}[t]{0.31\columnwidth}
\centering
\includegraphics[width=1.05in]{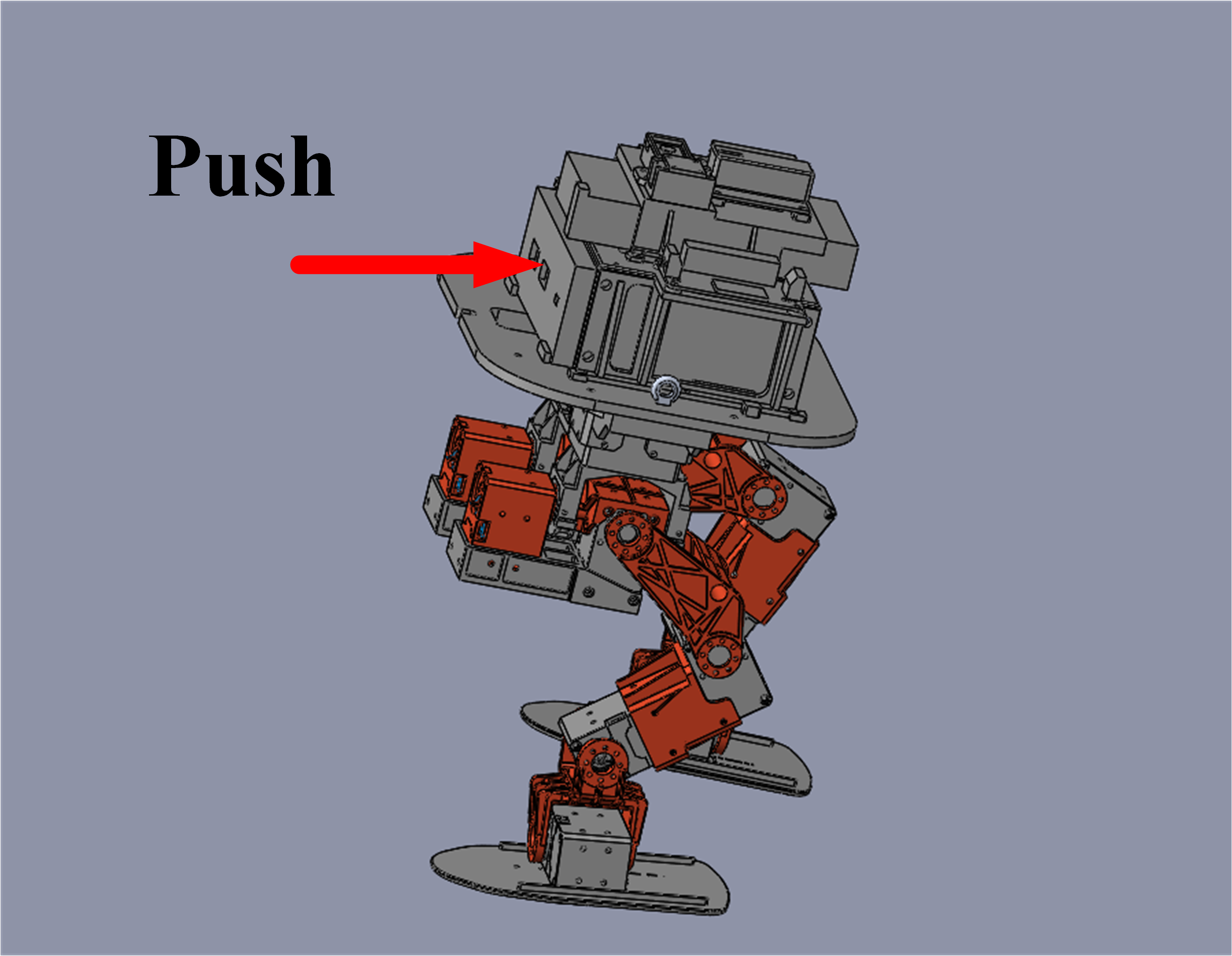}
%\caption{fig1}
\end{minipage}%
}%
\subfigure[obstacle]{
\begin{minipage}[t]{0.31\columnwidth}
\centering
\includegraphics[width=1.05in]{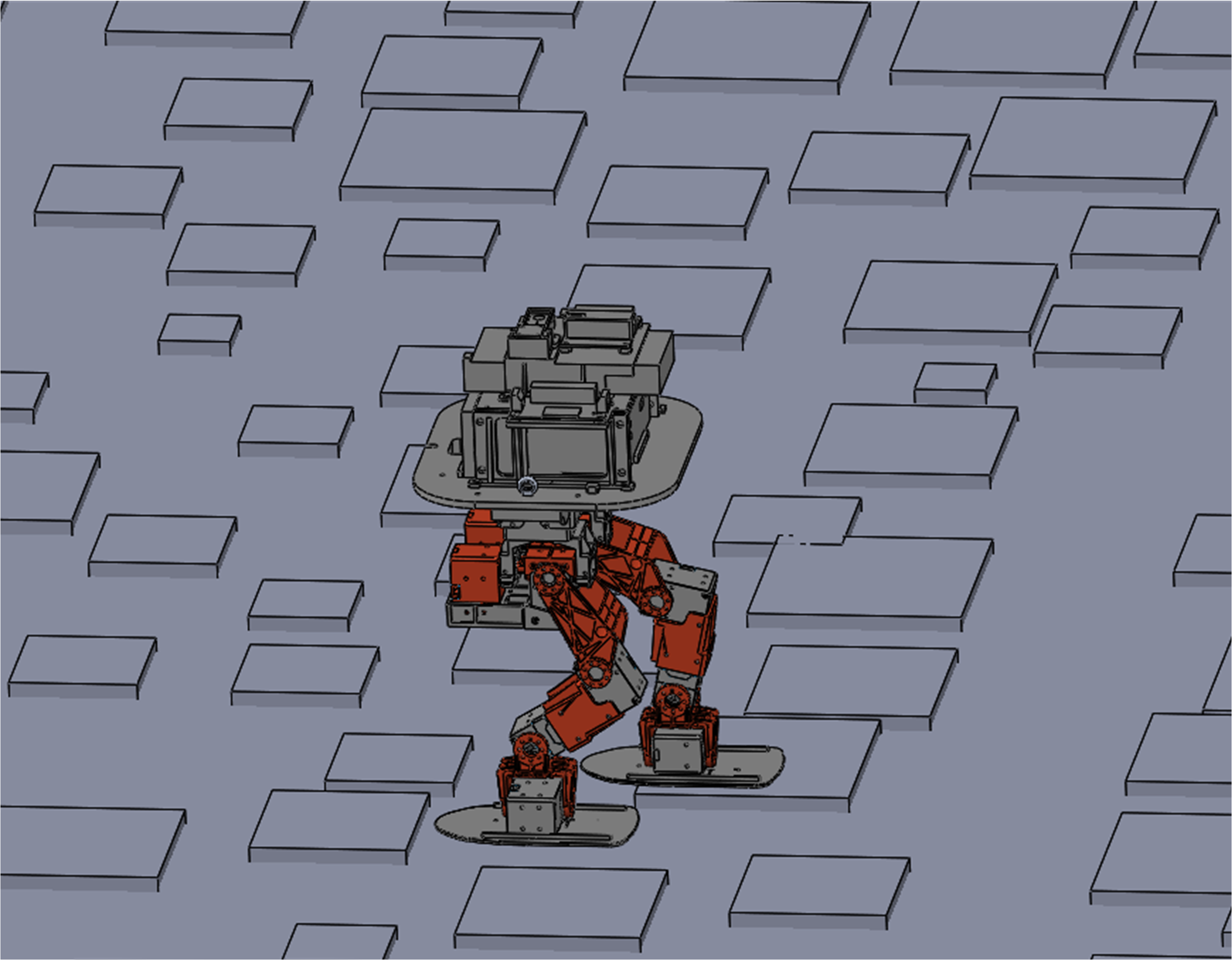}
%\caption{fig2}
\end{minipage}%
}%
\subfigure[slope]{
\begin{minipage}[t]{0.31\columnwidth}
\centering
\includegraphics[width=1.05in]{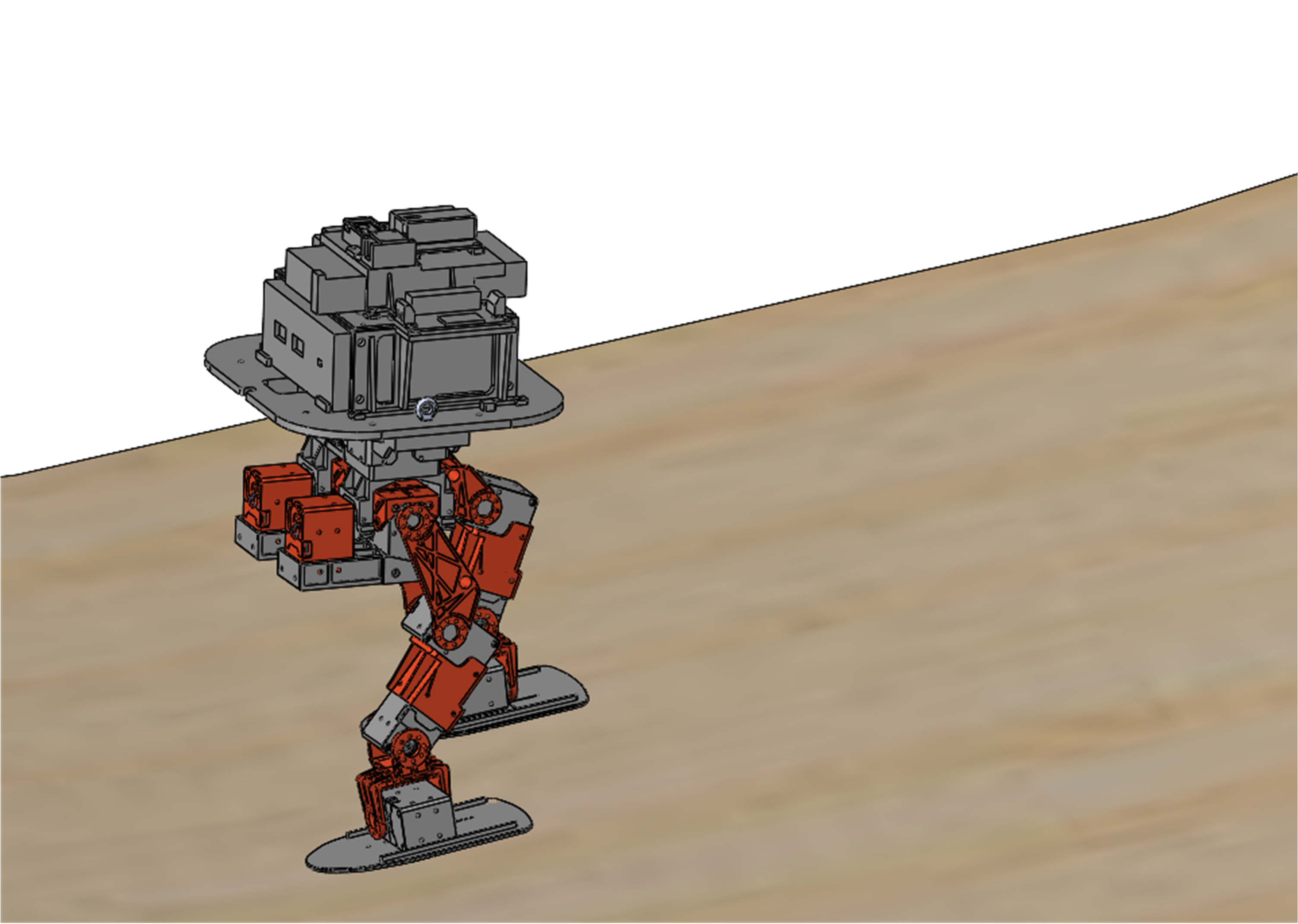}
%\caption{fig2}
\end{minipage}
}%
\centering
\caption{Illustrations of biped locomotion tasks: (a) walking under random disturbance; (b) walking cross obstacles; (c) walking on the slope.}
\label{fig:AIDA_tasks}
\vspace{-10pt}
\end{figure}

To verify the effectiveness of the proposed method, we conduct our simulator, named AIDA, to support development, training, and validation of biped locomotion. AIDA is built on Robot Operating System (ROS) framework \footnote{\url{https://www.ros.org/.}} and Gazebo simulation environment. We use Open Dynamics Engine (ODE) in our simulation, which is the default physical engine of Gazebo. The biped robot and the terrain are modelled with SolidWorks \footnote{\url{https://www.solidworks.com/}}, as shown in Figure \ref{fig:AIDA_tasks}. The robot contains 10 degrees of freedom (DoF), and each leg has 5 DoF. The center of mass of the pelvis is maintained at the center of the pelvis, which is similar to humans.

% We first describe the experiment setting and a biped locomotion benchmark in the section of ``Experiment setting".

% We train the agent on the first task and test its robustness on the remaining 3 tasks.

\subsection{AIDA robot}

\textbf{Simulation}. We built our AIDA bipedal robot on Robot Operating System (ROS) framework and Gazebo simulation environment. The biped robot is modelled with SolidWork, as shown in Fig. \ref{fig:robot}. It contains 10 degrees of freedom (DoF), and each leg has 5 DoF. We will release the AIDA simulator and the code of HDPG later.

\textbf{Physical robot}
The physical robot is equipped with a mini computer (Intel NUC8i7BEH), an IMU (YIS100-A-DK), 10 servo motors for joint control (DYNAMIXEL XH430-W210-T actuator) and a USB communication converter (U2D2).
At testing time, the trained policy is evaluated on NUC in real time. The policy network predicts appropriate controls based on only the current joint angles and IMU data. The U2D2 receives controls from NUC and transfers them to the motor controllers of the robot.

\subsection{Biped locomotion benchmark}\label{sec:benchmark}

To make comprehensive evaluation of the proposed method, we further define a biped locomotion benchmark that contains 3 walking tasks, which simulate 3 corresponding real-world challenging problems: walking under random disturbances, walking cross obstacle, and walking on the slope. The details of 3 tasks are described as follows:
\begin{itemize}
    \item {Walking under random disturbance: The robot walks under disturbances from different directions, as shown in Figure \ref{fig:AIDA_tasks}(a). The disturbance range is from 6N to 14N, and the duration is 0.2s. The directions contain forward, backward and sideward (either on the left or right side). We measure the success rate that the robot recovers from push disturbances.}

    \item {Walking cross obstacles: We randomly placed static obstacles of different heights on the ground, as shown in Fig. \ref{fig:AIDA_tasks}(b). We separately measure the success rate of the robot crossing obstacles of different heights.}

    \item {Walking on the slope: We build a slope with different angles to evaluate the terrain adaptability of the robot, as shown in Figure \ref{fig:AIDA_tasks}(c).We separately measure the success rate of the robot climbing slopes of different angles.}
\end{itemize}

%\begin{figure}[]
%\centering
%\includegraphics[width=0.65\columnwidth]{./figure/mhcritic.png}
%\caption{Multi-head critic architecture. The multi-head critic learns multiple values, each value corresponding one reward component. Different $Q$-values share the low-level layers parameters.}
%\label{fig:MHC}
%\vspace{-5pt}
%\end{figure}

% \begin{figure*}[t]
% \centering
% \includegraphics[width=0.9\textwidth]{./figure/training_curve_total.png} % Reduce the figure size so that it is slightly narrower than the column. Don't use precise values for figure width.This setup will avoid overfull boxes.
% \caption{Comparison of DDPG, MHDDPG and HDPG training curves. (a) is the Comparison of DDPG, MHDDPG and HDPG. Our HDPG achieves a much higher reward than MHDDPG and DDPG. It performs the highest learning efficiency at the same time. (b) is the training curve of each reward of MHDDPG. (c) is the training curve of each reward of HDPG.}
% \label{fig:reward_curve}
% \vspace{-10pt}
% \end{figure*}

\begin{figure*}[t]
\centering
\subfigure{
\begin{minipage}[t]{0.33\textwidth}
\centering
\includegraphics[width=1.73in]{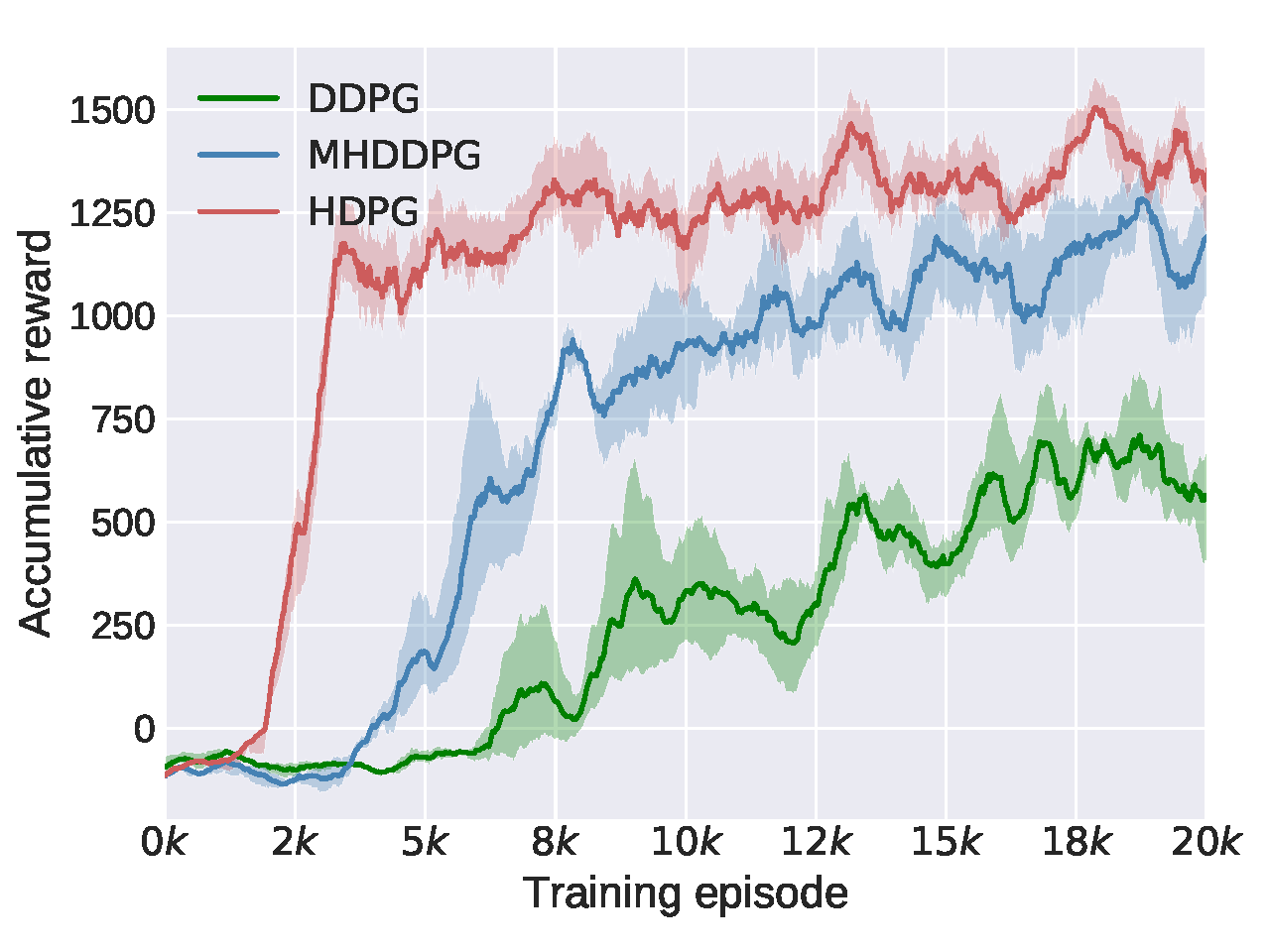}
\end{minipage}%
}%
\subfigure{
\begin{minipage}[t]{0.33\textwidth}
\centering
\includegraphics[width=1.73in]{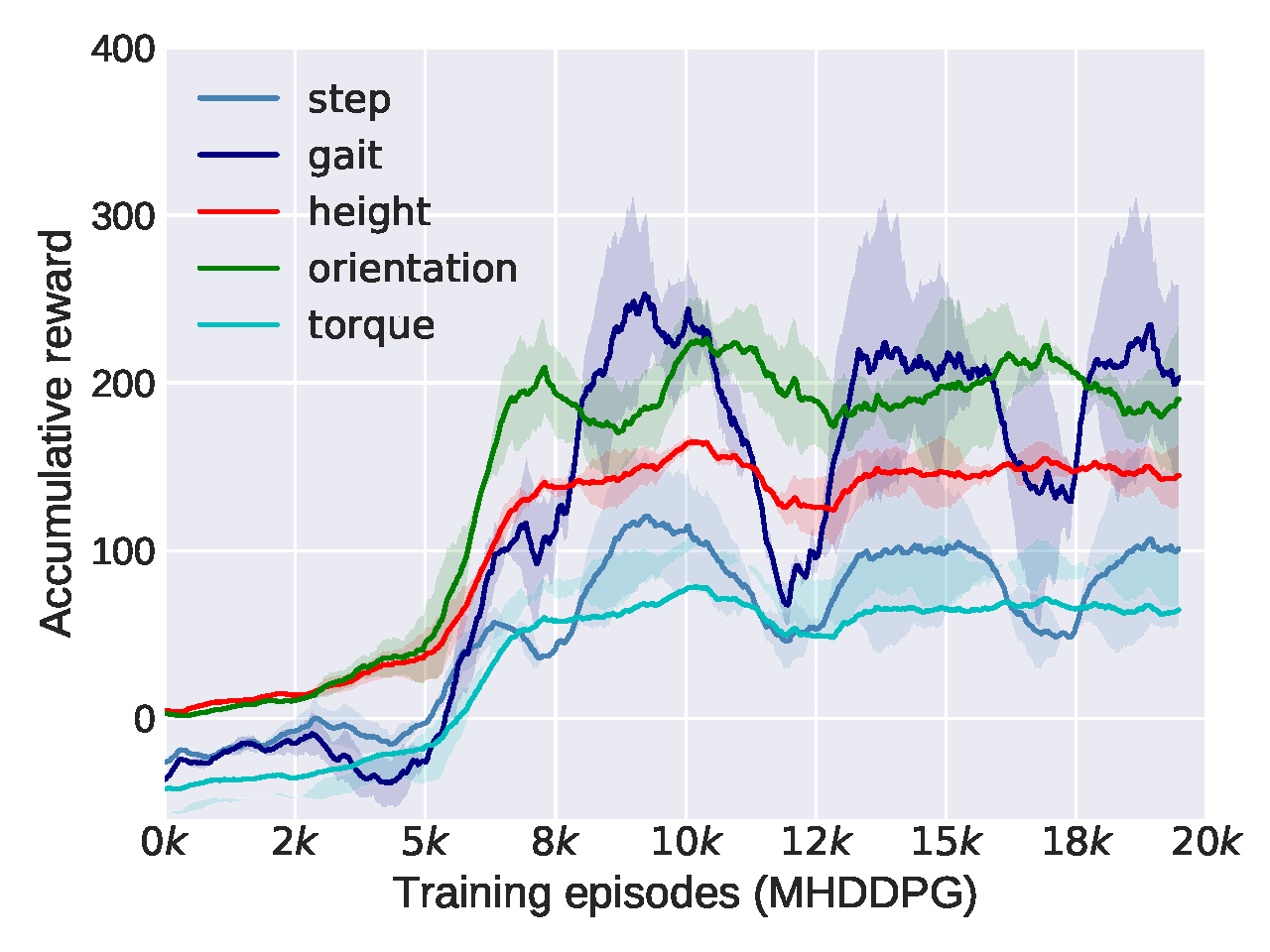}
\end{minipage}%
}%
\subfigure{
\begin{minipage}[t]{0.33\textwidth}
\centering
\includegraphics[width=1.73in]{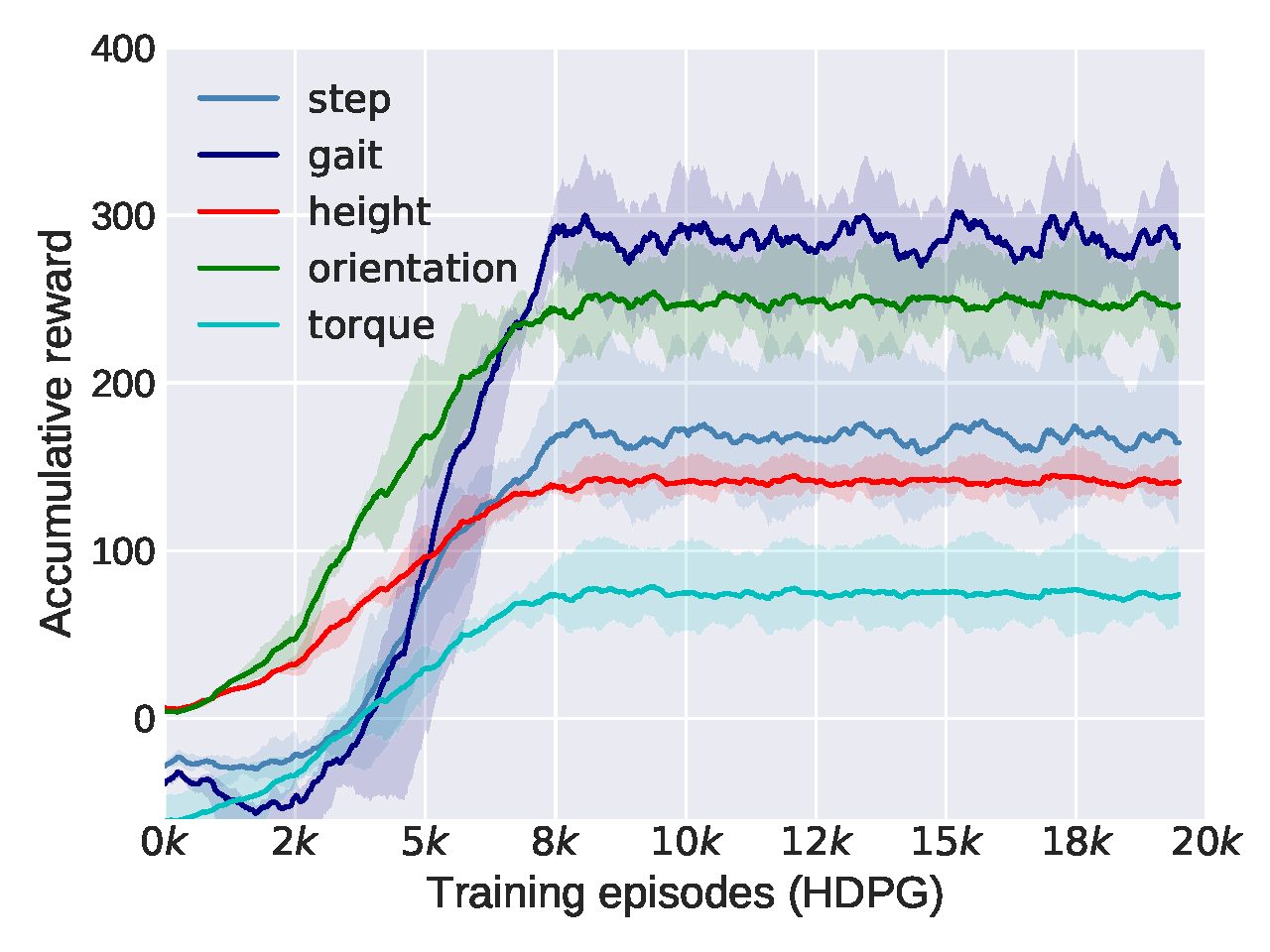}
\end{minipage}%
}%
\centering
\caption{Comparison of DDPG, MHDDPG and HDPG training curves. \textbf{Left}: The comparison of DDPG, MHDDPG and HDPG. Our HDPG achieves a much higher reward than MHDDPG and DDPG. It performs the highest learning efficiency at the same time. \textbf{Middle}: The training curve of each reward of MHDDPG. \textbf{Right}: The training curve of each reward of HDPG.}
\label{fig:reward_curve}
\vspace{-10pt}
\end{figure*}

\begin{figure*}[t]
\centering
\subfigure{
\begin{minipage}[t]{0.33\textwidth}
\centering
\includegraphics[width=1.8in, trim=8 25 8 1]{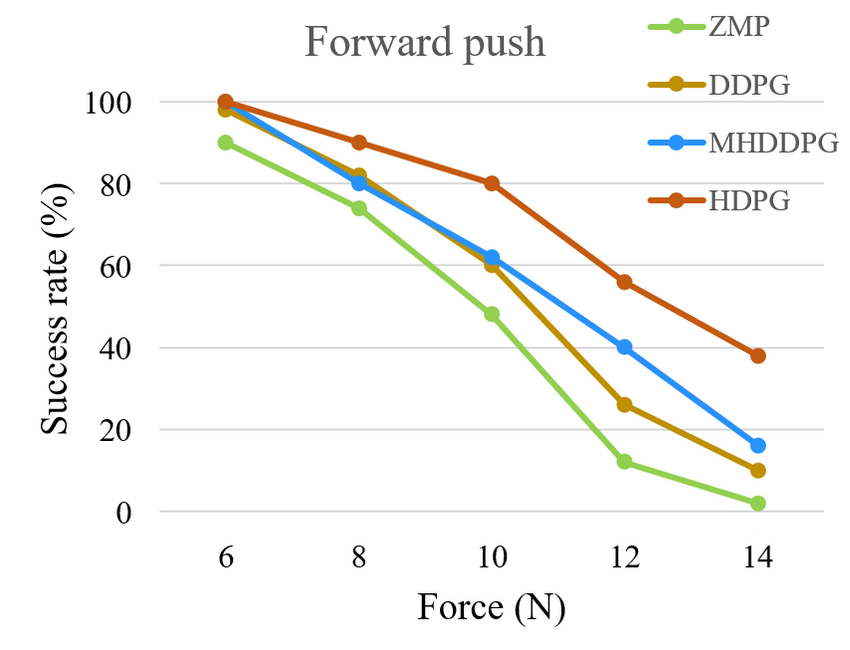}
\end{minipage}%
}%
\subfigure{
\begin{minipage}[t]{0.33\textwidth}
\centering
\includegraphics[width=1.8in, trim=8 25 8 1]{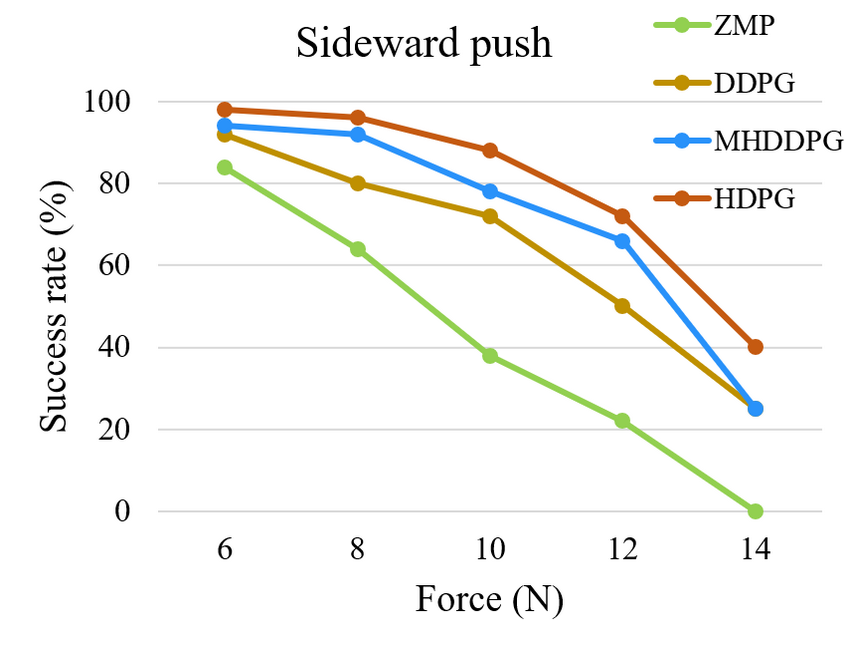}
\end{minipage}%
}%
\subfigure{
\begin{minipage}[t]{0.33\textwidth}
\centering
\includegraphics[width=1.8in, trim=8 25 8 1]{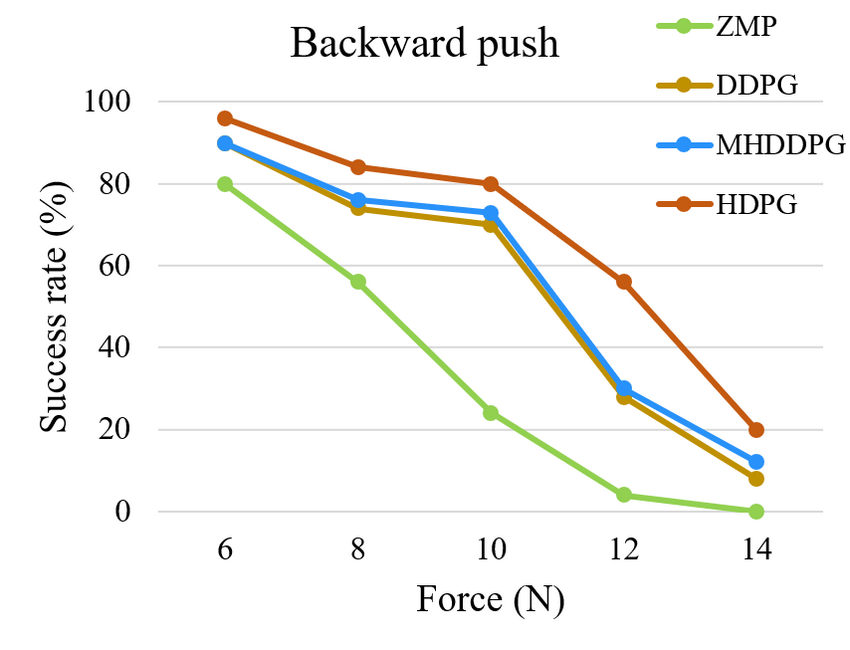}
\end{minipage}%
}%
\centering
\caption{Comparison of the success rate in ``walking under random disturbance" task. The robot is commanded to walk forward, and random push force will be applied to the robot's pelvis from different directions. We conducted 100 trials for each test. A trial is considered successful if the robot recovers stable walking from push disturbance.}
\label{fig:push}
\vspace{-10pt}
\end{figure*}

\subsection{Implementation details} \label{sec:implementation}

In this section, the implementation details of HDPG will be elaborated. The robot state $s_t$, as shown in Figure \ref{framework}, contains the IMU data and the angular position of each joint. In order to extract temporal features, 3 state frames $(s_{t-2},s_{t-1},s_t)$ are concatenated and fed into the policy network to predict an action. The combined states yield 69-dimensional state space (23D for each state). The predicted action is the angular position of each joint, which is a 10-dimensional vector.

% The computer used to perform all experiments equipped with an Intel i7-9700K CPU and a GeForce RTX 2070 GPU. It takes nearly 50 hours to train HDPG for 20k episodes. For DDPG and MHDDPG, it takes about 45h and 48h, respectively. As shown in Fig 3(a), the intermediate HDPG model trained with 8k episodes performs well enough and requires about 20 hours of training.

\textbf{Actor}. The actor network consists of 3 fully connected layers. The concatenating state $(s_{t-2},s_{t-1},s_t)$ passes through 3 fully connected layers of size (69,128), (128,256) and (256,10) to predict an action.

\textbf{Multi-head critic}. Multi-head critic aims to learn a separate value function through a multi-head network, which is similar to HRA \cite{van2017hybrid}. The action $a_{t}$ and the state $s_t$ are passes through one fully connected layer of size (10,128) and (69, 128), respectively. They are then concatenated and pass through two fully connected layers of size (256,256) and (256,6) to produce predicted 6 $Q$-values, each value corresponding one reward component, as shown in Fig. \ref{framework}.

\textbf{Training}. In the training process, the discount factor is set as 0.99. The learning rate of actor network and critic network are set as $1e^{-5}$ and $1e^{-4}$, respectively. And the training batch size is set as 64. The priority-based dynamic weight for hybrid policy gradient is updated every $T=20$ episodes. All training uses Adam optimizer \cite{KingmaB14}

\textbf{Reward normalization}. The weights of torque reward and step reward are referred to Walker2d of OpenAI Gym \cite{brockman2016openai}, where $w_1$ and $w_3$ are set as 1.0 and 0.001, respectively. The other weights (e.g., $w_2$, $w_4$, $w_5$) are set as 1.0.
We use simple max-min normalization for each component. To clarify how to normalize rewards, gait reward is given as an example. Firstly, the maximum achievable gait length $d_{max}^g$ is calculated according to the physical limitations. So the maximum and minimum gait rewards are $r_{max}^g=w_1d_{max}^g$ and $r_{min}^g=-w_1d_{max}^g$, respectively (according to Eq. \eqref{gait_reward}). Hence, any gait reward $r^g$ can be normalized to (0,1) by $(r^g - r_{min}^g) / (r_{max}^g - r_{min}^g)$. It is the same for other rewards.

%\begin{figure*}[t]
%\centering
%\subfigure[HalfCheetah-v3]{
%\begin{minipage}[t]{0.33\linewidth}
%\centering
%\includegraphics[width=1.9in]{./figure/cheetah.png}
%%\caption{fig1}
%\end{minipage}%
%}%
%\subfigure[Hopper-v3]{
%\begin{minipage}[t]{0.33\linewidth}
%\centering
%\includegraphics[width=1.9in]{./figure/hopper.png}
%%\caption{fig2}
%\end{minipage}%
%}%
%\subfigure[Walker2d-v3]{
%\begin{minipage}[t]{0.33\linewidth}
%\centering
%\includegraphics[width=1.9in]{./figure/walker.png}
%%\caption{fig2}
%\end{minipage}
%}%
%\centering
%\caption{MuJoCo results: evaluation curves of DDPG\cite{LillicrapHPHETS15}, MHDDPG and HDPG. (a) HalfCheetach-v3, (b) Hopper-v3, and (c) Walkerd-v3. The solid line and shaded regions represent the mean and the standard deviation across 10 runs, respectively.}
%\label{fig:mujoco}
%\vspace{-10pt}
%\end{figure*}

% The actor network consists of 3 fully connected layers. The concatenating state $(s_{t-2},s_{t-1},s_t)$ passes through three fully connected layers of size (69,128), (128,256) and (256,10) to predict an action.

% The multi-head critic network is shown in \ref{fig:MHC}, the action $a_{t}$ and the state $s_t$ are passes through one fully connected layer of size (10,128) and (69, 128), respectively. They are then concatenated and pass through two fully connected layers of size (256,256) and (256,6) to produce predicted 6 $Q$-values, each value corresponding one reward component.

% In training process, the discount factor is set as 0.99. The learning rate of actor network is set as $1e^{-5}$ and the learning rate of critic network is set as $1e^{-4}$.

\section{Results}\label{sec:results}

The simulation results and sim-to-real results of biped walking experiments are illustrated in Sec. \ref{sec:simulation results} and Sec. \ref{sec:sim to real}. Experimental results from both simulation and transfer-to-real robots are presented in this video \footnote{\url{https://youtu.be/3jvysrbeHzQ}\label{video}}. We finally conduct the experiments on MuJoCo of OpenAI gym \cite{brockman2016openai} to further demonstrate the generalization of HDPG, as illustrated in Sec. \ref{sec:mujoco}.
\subsection{Simulation results}\label{sec:simulation results}
In this section, we will qualitatively and quantitatively analyze the performance of the proposed HDPG with ZMP\cite{kajita2014introduction} method and DDPG \cite{LillicrapHPHETS15}, which is the baseline of the proposed method.

\begin{itemize}
    \item ZMP: ZMP-based algorithm \cite{kajita2003biped, kajita2014introduction} is a kind of traditional classical control method. It has been proven effective in many bipedal walking tasks.
    \item DDPG: DDPG \cite{LillicrapHPHETS15} is a advanced actor-critic model-free RL algorithm. It is also considered as the basic model with single-head critic.

    \item MHDDPG: MHDDPG represents the DDPG \cite{LillicrapHPHETS15} model equipped the proposed multi-head critic. We keep the experimental settings and parameter settings consistent with DDPG.
    The only difference is that we learn hybrid policy gradients with multi-head critic, and each policy gradient is assigned with the same static weight to optimize the policy (e.g., $m_1=m2=...=m_K=1$).

    \item HDPG: HDPG is the complete version of the proposed method. It can be regarded as MHDDPG incorporate with dynamic weight for each policy gradient.

\end{itemize}

\begin{figure}[t]
\centering
\subfigure{
\begin{minipage}[t]{0.49\columnwidth}
\centering
\includegraphics[width=1.6in, trim=8 25 8 8]{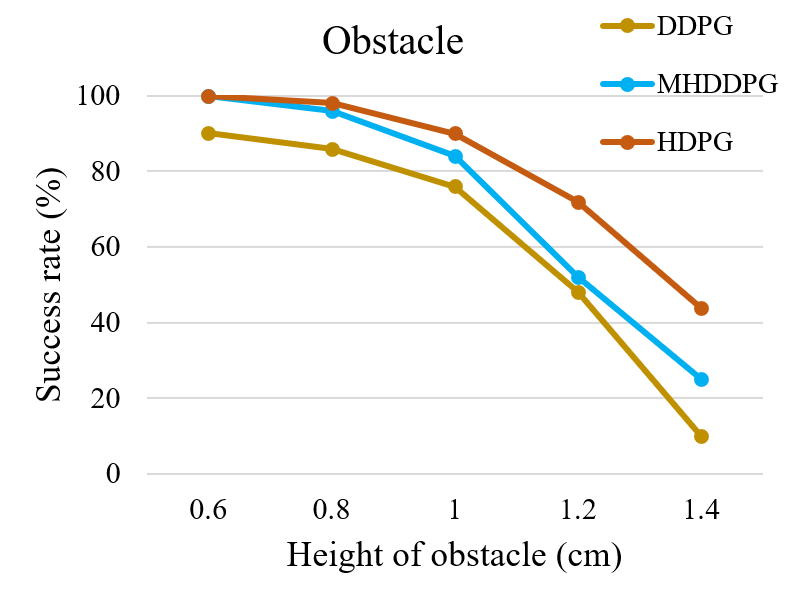}
\end{minipage}%
}%
\subfigure{
\begin{minipage}[t]{0.49\columnwidth}
\centering
\includegraphics[width=1.6in, trim=8 25 8 8]{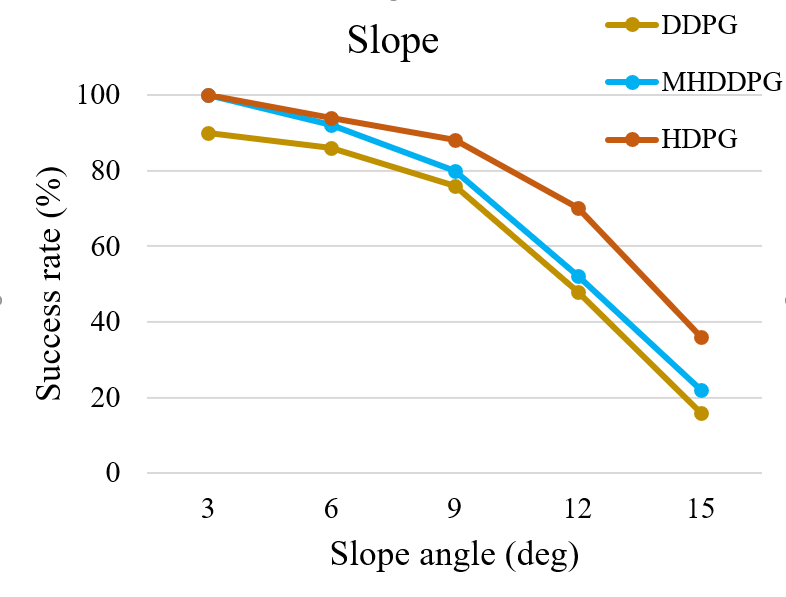}
\end{minipage}%
}%
\centering
\caption{Comparison of the success rate in ``walking cross obstacles" task and ``walking on the slope" task. A trial is considered successful if the robot is able to walk stable for 10 seconds. Left: The success rate of robot crossing obstacles with the height from range $(0.6,1.4)$(cm). Right: The success rate of robot walk on the slope with the angle from range $(3, 15)$(deg).}
\label{fig:os}
\vspace{-10pt}
\end{figure}

We train the bipedal robot with DDPG \cite{LillicrapHPHETS15}, MHDDPG and HDPG, respectively. The training curves are shown in Figure \ref{fig:reward_curve}(Left). The accumulative reward of HDPG starts to rise after training $2k$ episodes while MHDDPG and DDPG require $4k$ and $6k$ episodes, respectively. The performance of MHDDPG is much better than vanilla DDPG \cite{LillicrapHPHETS15}. Since vanilla DDPG simply sums the rewards up to a single value, it is unable to explicitly express the value of each component. This results in single Q-value not giving insight into factors contributing to policy, as mentioned in Sec. \ref{sec:method}. Some value functions are much easier to learn than others \cite{van2017hybrid}, then learning a separate value function is more effective. Obviously, the proposed HDPG outperforms DDPG and MHDDPG. With the utility of dynamic weight, HDPG can adjust the learning priorities of each gradient component and learn in a more efficient way.

% \begin{figure*}[t]
% \centering
% \subfigure{
% \begin{minipage}[t]{0.49\textwidth}
% \centering
% \includegraphics[width=2.1in]{./figure/mhddpg_neurips.png}
% \end{minipage}%
% }%
% \subfigure{
% \begin{minipage}[t]{0.49\textwidth}
% \centering
% \includegraphics[width=2.1in]{./figure/hdpg_neurips.png}
% \end{minipage}%
% }%
% \centering
% \caption{Training curve of each reward component. Left: the training process of MHDDPG. Right: the training process of HDPG.}
% \label{fig_each_reward}
% \end{figure*}

% \begin{figure}[t]
% \centering
% \includegraphics[width=0.45\columnwidth]{./figure/robot.png} % Reduce the figure size so that it is slightly narrower than the column. Don't use precise values for figure width.This setup will avoid overfull boxes.
% \caption{Left: The simulation bipedal robot conduct in Gazebo. Right: The physical bipedal robot AIDA. Each leg has 5 degrees of freedom, and the IMU is assembled in the pelvis, as indicated in the picture.}
% \label{fig:robot}
% \vspace{-10pt}
% \end{figure}

\begin{figure*}[t]
\vspace{10pt}
\centering
\includegraphics[width=0.93\textwidth, trim=8 25 8 8]{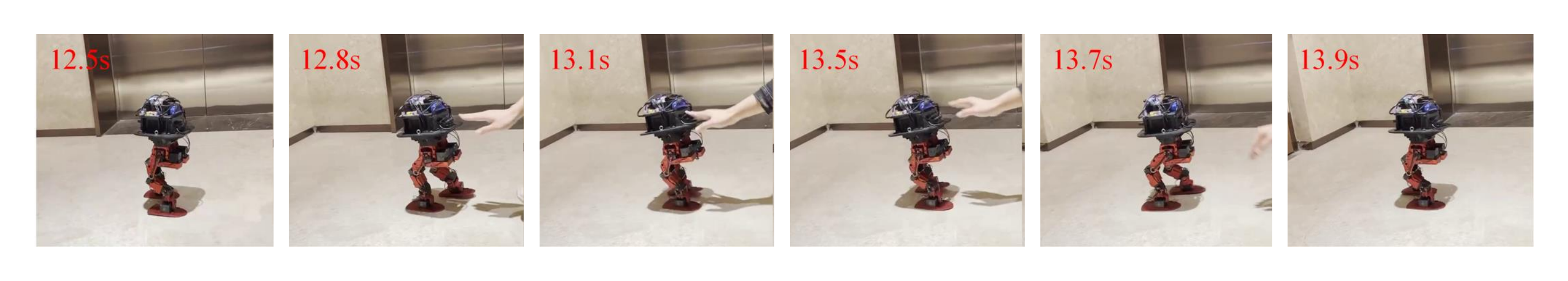} % Reduce the figure size so that it is slightly narrower than the column. Don't use precise values for figure width.This setup will avoid overfull boxes.
\caption{Simulation to real experiment. The HDPG policy trained with dynamics randomization in simulation is successfully transferred to physical robot. The bipedal robot is able to recover stable walking from the forward push force. Please see the YouTube link for a video version.}
\label{fig:real}
\vspace{-10pt}
\end{figure*}

\subsubsection{Walking under random disturbance}
To evaluate the robustness of the above several methods, the robot is asked to walk under various push disturbances from different directions, as mentioned in Sec. \ref{sec:benchmark} The success rate results are shown in Fig \ref{fig:push}. For 6N disturbances from 3 directions, all algorithms show good resilience. In 8N disturbance 8N from sideward tasks, only HDPG can maintain a success rate of over 90\%. In comparison, the success rate of DDPG and MHDDPG dropped sharply to about 80\%.
As the disturbance increases, the success rate of the ZMP-based method drops the fastest and exhibits the worst robustness. For example, when the disturbance increases to 10N, the success rate of ZMP in 3 directions is only 38\%, 38\% and 24\% respectively. While the other RL-based methods can maintain a success rate of over 50\%. As the disturbance increases to 12N and 14N, HDPG shows greater stability. It is at least about 10\% higher than other methods in forward tasks. In general, the proposed MHDDPG and HDPG outperform DDPG on all walking under random disturbance tasks, and HDPG achieves the best performance.

\subsubsection{walking cross obstacle}

We simulate the obstacle on the ground to further evaluate the robustness of different algorithms. In our experiment, obstacles are randomly placed on the ground. The success rate curve is shown on the left of Fig. \ref{fig:os}. Since ZMP agent can hardly walk cross obstacles, no comparison is made here. RL-based methods show better adaptability. In this task, HDPG performs the best robustness in crossing obstacles of all heights, while MHDDPG achieves the second performance. For 0.6cm obstacles, HDPG and MHDDPG are able to maintain a 100\% success rate of recovery. For 1.2cm obstacles, only HDPG achieves a success rate over 60\%, while MHDDPG and DDPG  50\% drop to about 50\%. For 1.4cm obstacles, although HDPG has the best performance, it is only about 40\%. This is due to the limitation of the height of the robot's leg lift.

\subsubsection{Walking on the slope}

Walking on the undulating terrain is another challenging task for a biped robot. We test the success rate of walking on the slope for 10 seconds, as shown in the right of Fig. \ref{fig:os}. The performance of DDPG and MHDDPG is close, which means that multi-head critic doesn¡¯t have a strong effect in this task. HDPG also has the best performance. When the slope angle is 12 degrees, HDPG shows more robust performance, and the success rate is nearly 20\% higher than the other two algorithms.

\subsubsection{The effectiveness of dynamic weight}
To further reveal the effectiveness of dynamic weight, we recorded the cumulative reward of each reward during training, as shown in Fig. \ref{fig:reward_curve} (Middle and Right). In order to observe the training characteristic of torque reward, we calculate the average torque reward for each time step and scale it up so it can be visible compared to other rewards.
For MHDDPG, height reward, torque reward and orientation reward learning are relatively stable. However, the step reward and gait reward are very unstable, which indicates that there is dependency between different reward components, and different components will influence each other during training. On the contrary, HDPG exhibits stable training of all 5 components, as shown in the Fig. \ref{fig:reward_curve}(Right). It is worth noting that the height reward and orientation reward first rises at about $2k$ episodes, while the gait reward and step reward then rise at about $6k$ episodes. This means that the robot first learns how to keep body balance, and then learns to walk forward. It implies that priority weight allows the robot to learn different components in a more orderly manner.

\subsection{Sim-to-real results}\label{sec:sim to real}

The policies trained with dynamics randomization in simulation are successfully transferred to physical robot. For dynamics randomization, we only randomize the pelvis center of mass, which is introduced in \cite{siekmann2020learning}. As shown in Fig. \ref{fig:real}, the biped robot was disturbed by push force at $13.1s$. Subsequently, The pelvis tilted at $13.5s$. As the policy adjusts the actions, the robot recovers to stable gait at $13.9s$. This demonstrates that the policy trained with HDPG is robust enough to handle the shift from simulation to reality.

\begin{table}[]
  \small
  \centering
  \caption{Performance on MuJoCo tasks. The results show the mean and standard deviation across 10 runs.}
    \resizebox{\linewidth}{!}{
    \begin{tabular}{l|ccc}
    \toprule
    Methods & Hopper-v3 & Walker2d-v3 & HalfCheetah-v3 \\
    \midrule
    DDPG & 3476.5$\pm$98.0 & 2398.5$\pm$1169.7 & 12422.9$\pm$620.4 \\
    MHDDPG & 3539.3$\pm$23.9 & 3162.0$\pm$1263.8 & 12748.9$\pm$501.8 \\
    HDPG & \textbf{3600.4$\pm$30.5} & 3217.5$\pm$1144.8 & \textbf{12895.3$\pm$528.1} \\
    \midrule
    PPO & 2609.3$\pm$700.8 & 3588.5$\pm$756.6 & 5783.9$\pm$1244.0 \\
    HD-PPO & 3263.5$\pm$367.8 & \textbf{4688.0$\pm$220.1} & 6165.2$\pm$150.4 \\
    \bottomrule
    \end{tabular}%
    }
  \label{tab:mujoco_results}%
\vspace{-10pt}
\end{table}%

\subsection{MuJoCo results}\label{sec:mujoco}

To further verify the generalization of the proposed method, we evaluate HDPG on 3 locomotion tasks on MuJoCo continuous control environment of OpenAI Gym \cite{brockman2016openai}, i.e., Walker2d, HalfCheetah and Hopper. We conduct MuJoCo experiment using the publicly released implementation repository\footnote{\url{https://github.com/thu-ml/tianshou}\label{footnote:tianshou}} as baseline.
% In addition, we incorporate the hybrid and dynamic policy gradient idea into proximal policy optimization algorithms (PPO) \citet{schulman2017proximal}, which is an advanced on-policy algorithm in the policy-gradient family. This variant of PPO is denoted as HD-PPO in the following discussion.
Following the same experiment settings, we compare the performances of DDPG, MHDDPG, and HDPG. As shown in the Table \ref{tab:mujoco_results}, MHDDPG and HDPG outperform DDPG baselines on all tasks within 1M environment steps, and HDPG achieves the best performance on all tasks. It implies that the proposed ``learning a separate value function" and ``dynamic policy gradient" can be applied to more general continuous control tasks. Furthermore, we equip HDPG with proximal policy optimization algorithms (PPO) \cite{schulman2017proximal} to verify the generalization of HDPG, which is denoted as HD-PPO. The best results are highlighted in bold. Hopper and Walker tasks contain 3 reward components, while HalfCheetah contains 2 reward components. Obviously, hybrid and dynamic policy gradient method is able to improve the performance of DDPG and PPO on all tasks. It is worth noting that HD-PPO has achieved a breakthrough of more than 25\% on Hopper and Walker tasks, while it has nearly increased by about 6.6\% on HalfCheetah task. This demonstrates that the proposed method obtains relatively unobvious improvement on tasks with simple rewards, such as HalfCheetah, where there are only 2 components in its reward function. While the improvement is obvious in Hopper and Walker tasks, where there are more components in its reward function.

\section{Conclusion}

In this work, we propose a reward-adaptive reinforcement learning method for bipedal locomotion tasks, also called hybrid and dynamic policy gradient (HDPG) optimization. It decomposes the commonly adopted holistic polynomial reward function and introduces priority weights to enable the agent to adaptively learn each reward component. Experimental evaluation illustrates the effectiveness of HDPG framework by showing better performance on Gazebo simulation in perturbation walking challenges, walking cross obstacles challenges and walking on undulating terrain challenges. With dynamics randomization, the policy trained in simulation is successfully transferred to a physical robot. In addition, we further verify the generalization of HDPG on 3 MuJoCo tasks. Our future work is to transfer the policy trained in simulation into real physical robots in walking across obstacle challenges and undulating terrain challenges.

\section{Acknowledgment}
The authors would like to thank Jiang Su, Zhihong Zhang and Dong Zhao for fruitful discussions and their valuable comments on this paper.

\bibliographystyle{IEEEtran}
%\bibliography{IEEEabrv,myref}
\bibliography{reference}

\end{document}